# SwarmCloak: Landing of Two Micro-Quadrotors on Human Hands Using Wearable Tactile Interface Driven by Light Intensity

Evgeny Tsykunov[1], Ruslan Agishev[1], Roman Ibrahimov[1], Taha Moriyama[2], Luiza Labazanova[1], Hiroyuki Kajimoto[2], and Dzmitry Tsetserukou[1]

## ABSTRACT

For the human operator, it is often easier and faster to catch a small size quadrotor right in the midair instead of landing it on a surface. However, interaction strategies for such cases have not yet been considered properly, especially when more than one drone has to be landed at the same time. In this paper, we propose a novel interaction strategy to land multiple robots on the human hands using vibrotactile feedback. We developed a wearable tactile display that is activated by the intensity of light emitted from an LED ring on the bottom of the quadcopter. We conducted experiments, where participants were asked to adjust the position of the palm to land one or two vertically-descending drones with different landing speeds, by having only visual feedback, only tactile feedback or visual-tactile feedback. We conducted statistical analysis of the drone landing positions, landing pad and human head trajectories. Two-way ANOVA showed a statistically significant difference between the feedback conditions. Experimental analysis proved that with an increasing number of drones, tactile feedback plays a more important role in accurate hand positioning and operator's convenience. The most precise landing of one and two drones was achieved with the combination of tactile and visual feedback.

## 1 Introduction

While large drones [1,2] are capable of lifting high-performance vision and processing systems for autonomous navigation and landing, the swarm of micro-quadrotors cannot process the visual data autonomously. The actual flight of drones often does not require high accuracy of positioning, and, therefore, autonomous flight can be easily accomplished with limited sensing capabilities, such as GPS. However, takeoff and landing operations often require an accurate positioning system which could be a problem for micro-quadrotors. Hence, the human could supplement these challenging swarm operations. For the human operator, it is often easier and faster to catch a small size quadrotor right in the midair instead of landing it on a surface in autonomous mode. The reasons for this could be multiple. For the outdoor applications, the landing surface is usually uneven and dusty, which could lead to a crash of the swarm. Even when the landing spots (helipads) are provided, autonomous landing is not always the best solution due

1 - authors are with the Intelligent Space Robotics Laboratory, Skolkovo Institute of Science and Technology, Moscow, Russian Federation (email: {Evgeny.Tsykunov, Ruslan.Agishev, Roman.Ibrahimov Luiza.Labazanova, D.Tsetserukou}@skoltech.ru).
2 - authors are with the University of ElectroCommunications, Tokyo, Japan (e-mail: {moriyama, kajimoto }@kaji-lab.jp).

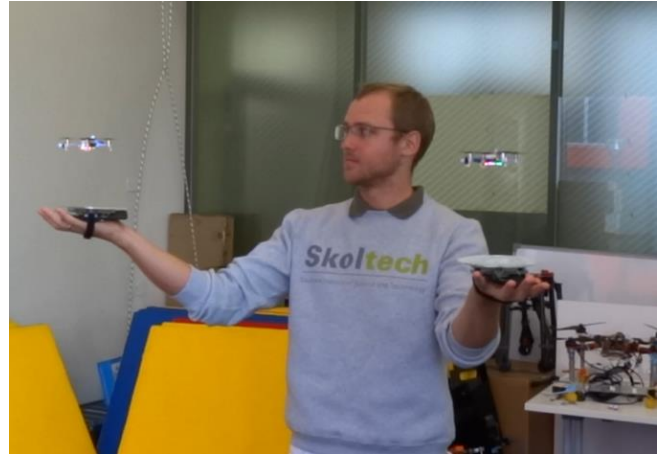

Fig. 1. A user lands two drones on the landing pads.

to position estimation errors, robustness, or high cost of a positioning system. On the other hand, the human can try to catch the drones from the formation while the fleet is descending. Nevertheless, this scenario could be dangerous both for the human and robots if the number of robots exceeds two. Considering well-developed human-robot and human-swarm interaction field [3], to our knowledge, up until now, there are no technologies and research on how to promptly deploy and land the swarm of drones using the human body. Hence, a robust interaction strategy has to be developed.

Haptic feedback for robot control has been widely investigated as reported in [4,5]. A single drone is used by Knierim et al. [6] for the real representation of digital content in virtual reality, providing the tactile sensation. An arm-worn tactile display for presentation of the collision of a single flying robot with walls was proposed in [7]. Vibrotactile signals improved users' awareness of the presence of obstacles. Aggravi et al. [8] developed a wearable haptic display capable of providing a wide range of sensations by skin stretch, pressure, and vibrotactile stimuli. The above-mentioned projects demonstrate that tactile feedback can be effectively applied for the control and interaction with drones.

In this paper, we propose a novel system SwarmCloak for drone deployment in mid-air. Wearable tactile display with a light sensor makes it possible to land the fleet of nano-quadrotors on the human hands (see Fig. 1). The developed technology is based on a hypothesis that tactile feedback could improve the accuracy of landing, and human convenience, especially in the cases when several drones are landing on the human limbs simultaneously.

We also consider the strengths of the SwarmCloak in comparison to the autonomous landing platforms, where the robust controller incorporating the accurate position information, could accomplish a precision landing. Autonomous landing requires a complex infrastructure which should include a position estimation system, e.g. motion capture system with infrared (IR) markers or regular cameras with visible markers, which has to track all landing pads with centimeter accuracy. Such positioning systems can be expensive or not reliable. Additionally, ground-based positioning systems are bulky. On the other hand, in the proposed approach, the formation only has to roughly estimate the position of the human, within reachable area by the human hands, to land vertically. We applied a Vicon motion capture system during the experiments to provide submillimeter accuracy of detection of drone and hand position during experiment.

## 2   Design of the Tactile Interface

The purpose of the designed tactile interface is to deliver the information about the position of the drone relevant to the landing pads to the operator. The vibration, which is activated by light, is proportional to the light intensity. If the drone is far away, no vibration occurs. While the drone approaching the human hand, the vibration intensity is gradually increasing. The location of the tactile stimulus reveals the location of the drone in a horizontal plane and stimulus intensity shows the distance to the robot in the vertical direction. The higher the intensity the closer drone to the user's skin.

The overall system consists of two landing pads (with light sensors and vibration motors) and two drones with LEDs on the bottom. The single sensor-tactor unit, shown in Fig. 2, is based on HALUX technology [9] and comprises a linear resonant actuator (LRA) (LD14- 002, Nidec Copal Corporation), a photo-transistor (PT19-21C, Everlight Electronics CO., Ltd.), and an oscillation circuit for LRA. LRA was selected for its fast response of less than 20 ms. Optimal sensitivity of the skin is achieved at frequencies between 150 and 300 Hz, according to research findings [10]. Meanwhile, the resonance frequency of the oscillation circuit with LRA is 150Hz. Therefore, the vibration frequency is set to 150Hz.

The amplitude of vibrations is modulated by the photo-transistor. During the landing stage, the distance $D$ between the drone and the landing pad is reducing. At the same time, the illuminance of the photo-transistor PT19-21C is increasing along with decreasing the illuminated area (LED viewing angle is fixed). Therefore, the illuminance is inversely proportional to the $D^2$. We set the relationship between the produced photocurrent and the vibration amplitude to be linear. As a result, when the drone is getting closer to the landing pad, the user experiences more intensive vibration. We keep the discussed vibration settings for all experiments which involve tactile feedback.

The model of the developed device along with main components is shown in Fig. 3. The electronic circuit of each sensor-tactor unit is placed in the plastic cover which has a hole above photo-transistor with a diameter of 3 mm for the light penetration (Fig. 3). The hole is of 10 mm deep, to protect the sensor from undesirable environmental lights and infrared emission of the motion capture system. The phototransistors are pointed upwards to detect the light emitted from the array of LEDs at the drone's belly. The landing plate (Fig. 3) is made of transparent acrylic material (diameter and thickness of plate is 160 mm and 3 mm, respectively).

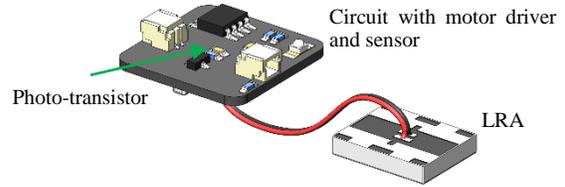

Fig. 2. 3D model of the sensor-vibrator unit.

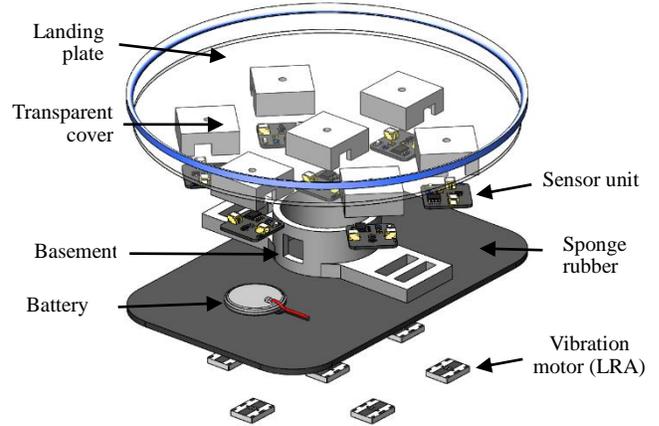

Fig. 3. Layout of the landing pad.

Palm has a flat and wide contact area with high tactile resolution [11]. We have placed seven LRAs on the human palm to achieve easily distinguishable stimulus (the distance between neighbor LRAs is not less than 2 cm). Six units are uniformly distributed on the circle of radius 40 mm and one unit is placed in the center. All of the units are mounted on the back of the transparent landing plate, with photo-transistor pointed upwards. The operational mode of each unit is the same.

The LRAs are fixed to a thin sponge rubber pad that is placed directly on the palm. To provide an explicit sensation of the drone position above the palm, the arrangement for the vibrators is selected in a way that it replicates the position of the sensors. The wearable device is attached to human hand by Velcro tape.

For the experiments, we used two Crazyflie 2.0 quadrotors. Small size (9 cm$^2$) and light weight (27 grams) secure safety, which is crucial for applications involving human-robot physical interaction. To protect user's eyes from false movement of drones subjects have worn safety glasses during the experiment. Vicon motion capture system with 12 IR cameras covering 5 m × 5 m × 5 m space tracked the quadrotors, landing pads, and the human hands. We used the Robot Operating System (ROS) Kinetic framework to run the custom software and ROS stack for Crazyflie 2.0. Sensors of the landing pad are sensitive to the infrared spectrum, for that reason we have decreased the intensity of IR strobe from the motion capture cameras.

Airflow from the landing quadrotor could provide strong tactile cues, which may actually be used as a source of additional information about the position of drones. As far as in the experiments we aimed to investigate only vibrotactile feedback, the effect of tactile cues to the hand, which is caused by the airflow, was canceled by the increased size of landing plate with additional cardboard. The size of the cardboard was 300 x 300 mm. It was employed only for the experiments and in the case of real-life applications, the cardboard is not needed.

## 3 Experiments

Seven right-handed users (six males and one female, 24 to 41 years old) took part in the experiments in which they landed one or two drones on the palms. In particular, the subjects were asked to adjust the position of the landing pads so that each descending drone could land in the middle of the corresponding pad (Fig. 1). There were three feedback conditions: only visual feedback, only tactile feedback, and tactile-visual feedback. The protocol of the experiment was approved by the Skolkovo Institute of Science and Technology review board and all participants gave informed consent.

### 3.1 Experimental Methods

For the user study, we used two landing pads, presented in Fig. 3, placed on the human palms. Two types of experiments were conducted. Seven people participated in the study and all seven subjects performed both experiments. During the experiment, the drones descended vertically keeping the same position in the XY plane. The goal of the subject was to adjust the position of the landing pads, in a way that the drones land on the center of the landing pads. In Experiment 1, subjects were asked to land one drone on the right hand. Experiment 2 was more complicated, as two drones were descending on both palms at the same time (distance between drones was 1 meter; therefore, it was possible to observe visually only one of them at a time. This fact forced subjects to move the head from side to side).

The basic guidance policy was proposed to the users. In each experiment, users were asked to adjust the position of the palm to land one or two drones as close to the center of the landing plate as possible. If the user feels that the drone is above the right side of the palm (with the help of vision or vibromotor activation placed on the right side of the palm), then he/she was supposed to move the hand to the right. After the experiments, users were asked about the applied strategy for a combination of different feedback types and the results are discussed below.

In both experiments, users were asked to land drones using one of three feedback conditions: only visual feedback (V), only tactile feedback (T) with closed eyes, or both visual and tactile feedback (VT). Users experienced the same tactile sensation for T and VT cases. Each feedback condition repeated 10 times in a random order (10 times for each of three conditions: V, T, and VT). As a result, in each experiment, one subject had 30 trials of landing. For specific feedback condition, landing speed varied in a random order, so that 5 times landing speed was slow (0.1 m/s), and 5 times it was fast (0.15 m/s). Hence, all users experienced 6 conditions (set of 3 feedback types and two landing speeds) with 5 trials for each condition.

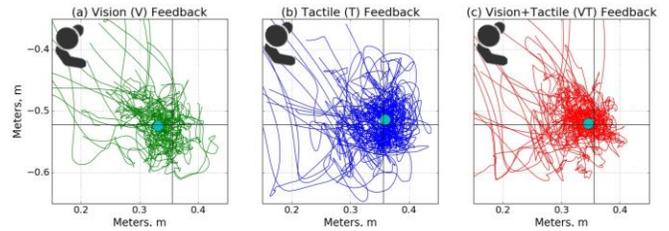

Fig. 4. Trajectories of all participant's right hand in Experiment 2 for slow drone landing in XY plane (for the right hand). Drone is landing on the intersection of black lines. Average position of the landing pad is shown with blue circle.

Before the experiments, users were asked to stand in the predefined spot and lower their hands. Users were not allowed to take steps while the drones were landing. Drones were placed in front of the users on the floor. They took off to the 2-meter height above the floor, then moved to the predefined position (approximately 0.5 meters in front of the human). Predefined positions were randomly selected within a range of 0.12 meters to prevent the learning of hand positioning. After that, the LED rings on the bottom of the quadrotors started to flash with constant light intensity in a visible spectrum and the drones started to descend vertically. Subjects were told that they were allowed to start to adjust the position of the landing pad when the LED ring was on. When the difference between the height of the drone legs and the landing plate was less than 5 mm, the motors of the corresponding drone shut down. Turning off the motors also helped preventing the drone drifting due to the aerodynamics of the ground effect. Ground effect was quite noticeable in the case of tactile feedback when users were not able to maintain the horizontal position of the landing plate visually. Slightly tilted plate led to the drone drifting and jumping in the direction of tilt during the last 5 mm of landing. The height of the contact point (when the drone is landed) can be selected by the subjects without any constraints. For Experiment 2, users were restricted to land both drones approximately at the same time, preventing sequential landing. After landing, drones were placed back on the floor and the process was repeated.

For both experiments, the training involved one fast and one slow landing for each feedback condition. During learning, users were able to get feedback about the distance between the center of the drone and the center of the landing plate by closely observing after landing. Each participant was wearing safety glasses.

Drones and landing plates were tracked by a Vicon motion capture system which recorded the position and orientation. For Experiment 2, we also asked participants to wear a cap which was tracked as well, for the analysis of the human head motion while catching both drones at the same time. Recording started after the drones initiated descending (after the LED ring started to flash) and stopped after the contact of a drone with landing plate.

### 3.2 Results and Discussions

*3.2.1 Trajectory analysis (of the landing pads and the user's head) during the landing stage.*

We analyzed the kinematic parameters and shape of the trajectories of human hands and human's head movement while landing drones. The landing stage begins when the drones start to descend and last until the drone actually touches the surface of the landing pad. For the analysis, we propose to consider the first four derivatives of the human position. Changes in the motion of parts of the human body could have a significant effect on the human experience. In general, humans are trying to minimize the changes in motion and the motion itself while doing different operations. Higher derivatives could have a strong effect on the human although human tolerance to snap and jerk are not well investigated. However, many designers of elevators and roller coaster rides prefer to limit these parameters [12]. The results are shown in Table 1 and Fig. 4.

Landing velocity affects the hand motion with V feedback for two drones (Experiment 2). When the drones descended faster, the human adjusted hand in a more aggressive way (snap increased by multiple times for fast landing speed (see Table 1, last column, Experiment 2, V rows)). Although this effect was not occurred when we added tactile sensation to vision in the VT case (see Table 1, last column, Experiment 2, VT rows)). This finding tells that tactile feedback helps to make human motion more smooth when we try to land multiple drones.

TABLE I. PARAMETERS OF HAND MOTION. <u>DURING</u> LANDING

| Feed-back type | Kinematic parameters, mean values, Slow / Fast landing | | | |
|---|---|---|---|---|
| | *Velocity, m/s* | *Acceleration, m/s$^2$* | *Jerk, m/s$^3$* | *Snap, m/s$^4$* |
| **Experiment 1. One drone. Right hand** | | | | |
| V | 0.026/0.025 | 0.20/0.22 | 12.9/12.7 | 1353/1306 |
| T | 0.029/0.043 | 0.21/0.30 | 11.8/15.0 | 1195/1570 |
| VT | 0.028/0.034 | 0.25/0.28 | 13.9/14.5 | 1488/1473 |
| **Experiment 2. Two drones. Left hand** | | | | |
| V | 0.025/0.027 | 0.22/0.25 | 10.4/18.2 | 984/3816 |
| T | 0.031/0.038 | 0.27/0.30 | 13.9/12.3 | 1702/1182 |
| VT | 0.023/0.032 | 0.22/0.26 | 10.3/12.4 | 935/1167 |
| **Experiment 2. Two drones. Right hand** | | | | |
| V | 0.028/0.027 | 0.23/0.27 | 11.8/21.5 | 1133/4712 |
| T | 0.033/0.044 | 0.28/0.37 | 12.4/14.9 | 1207/1415 |
| VT | 0.024/0.028 | 0.22/0.26 | 11.3/12.3 | 1061/1124 |

For one drone case (Experiment 1), with T feedback, participants demonstrated more active landing plate adjustment for a fast landing. This shows that the proposed device design could inform the users about the rate of change of the distance between the drone and the landing plate.

Most users in Experiment 2 demonstrated slightly more dynamic work with the right hand than with left hand having V feedback (snap is 20% higher). All participants are right-handed and could control the right hand more fast and precisely. But, again, this effect became negligible comparing with the VT case for the right and the left hand.

Fig. 4 presents the landing pad trajectories of the right hand of all users in Experiment 2 (in XY plane) during the landing stage. The intersection of black lines is the position of the landing drone, which is moving vertically. It is easy to notice that in V case the average position of the landing pad (showed with a blue circle) has an offset towards the location of human standing in the left upper corner. In contrast to that, in T case, the landing plate is moving below the drone without a noticeable offset. We averaged the distance in XY plane between the drone and the landing plate during the landing stage (measured before the drone touches the landing pad, for the slow landing on the right hand in Experiment 2): V: 24.2 mm, T: 8.6 mm, VT: 10.1 mm. Based on this evidence, we suggest that the tactile feedback helps align the position of the landing pad in such a way that the drone is located above the center of the pad during the landing.

One more finding is related to the motion pattern of the hand. Most subjects stated that it was easier to estimate the position of the drone based on the gradient of tactile sensation rather than when the vibration is always in the same palm spot. As a result, participants having T feedback always moved their palms from side to side (while landing). With V feedback, participants also adjusted the hand position all the time, trying to catch the drone with a smaller error. Trajectory analysis reveals that when visual feedback is presented (Fig. 4(a)), human mostly moves his/her hand along the line, which connects human and the drone. However, in trials with only tactile feedback (Fig. 4(b)), we see that hand motion is omnidirectional, which tells us that the users are exploring all space in a more uniform manner.

In Experiment 2 participants had to rotate its head fast to observe landing drones one by one (distance between drones is 1 meter). That is why we conducted the same trajectory analysis for the human head movement for the second experiment. The results are presented in Table 2 for V and VT conditions. Comparing V and VT cases in Table 2, it is easy to notice that VT feedback minimizes and smooths the human head motion. That means that VT requires to perform less locomotion.

TABLE II. HUMAN HEAD MOTION PARAMETERS, <u>DURING</u> LANDING.

| Feed-back type | Average kinematic parameters, Slow/Fast landing | | | |
|---|---|---|---|---|
| | *Vel., m/s* | *Accel., m/s$^2$* | *Jerk, m/s$^3$* | *Snap, m/s$^4$* |
| **Experiment 2. Two drones.** | | | | |
| V | 0.085/0.11 | 0.72/1.24 | 27.4/121 | 2676/35947 |
| VT | 0.080/0.087 | 0.60/0.69 | 22.1/26.5 | 2129/2543 |

Users reported that they switched their attention from one drone to the other when landing both. That is true for V, T and VT case, therefore, tactile feedback also requires individual attention, the same as vision. The most popular strategy for VT and two drones was to set one landing pad position with vision and then use tactile sensation to update the position of it, while the second landing pad was positioned with vision mainly.

*3.2.2 Landing position analysis.* An important metric for the experiments was the distance between the center of the drone and the center of the landing plate after landing. In the current paper, this distance is called displacement. The diameter of the landing plate was selected to be big enough so that in most experiment trials participants were able to land a drone on its surface.

First of all, to compare the effects of each condition (the combination of feedback type and landing speed) on the displacement, we used a within-subject statistical comparison. We performed a two-way ANOVA with repeated measures, in which

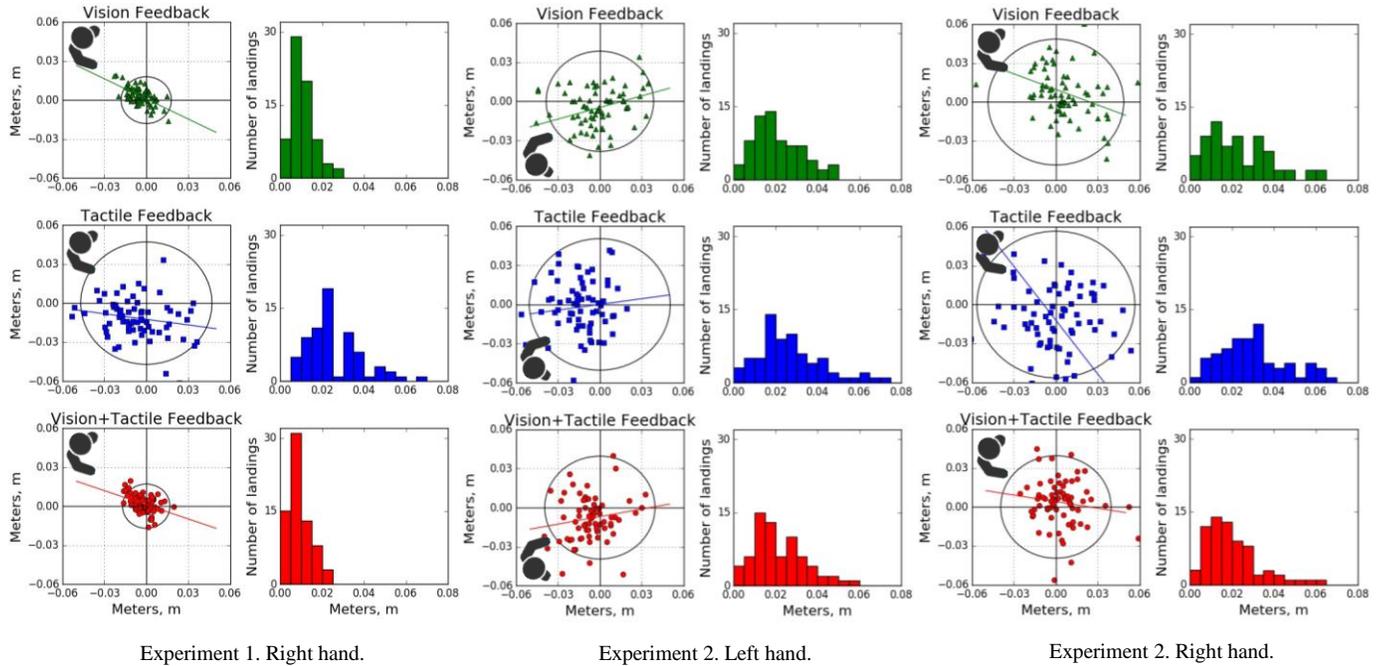

Experiment 1. Right hand.  Experiment 2. Left hand.  Experiment 2. Right hand.

Fig 5. Drone positions on the landing pad after landing. XY axis are crossing in the center of the landing plate. Histograms represent the distribution of the displacements. Circles represents the area that fits 90% of landings. Lines represent predictions of drone landing spot, based on linear regression model.

the dependent variable is displacement error, whilst the two factors are drone number and feedback/speed conditions. The level of significance was set to $\alpha = 0.05$. The analysis revealed statistically significant difference in all conditions (F(5, 170) = 9.459, p = 5.653 * $10^{-8}$). A number of drones do not affect the results significantly (F(5, 170) = 1.027, p = 0.404), thus, we can conclude that technology works similarly for landing one or two drones.

For the further displacement analysis, we used mean values of displacements and standard deviations, presented in Table 3 and paired t-test for different conditions. All displacement values of the drone after landing (with histogram), are plotted in Fig. 5.

The statistics of drone displacement changes drastically when the number of drones to land is changed from one to two (Table 3). V and VT cases revealed that the increase in the number of agents decreased the accuracy 2-3 times, although T condition performance remained the same. It can be concluded that the performance gap between visual (V) and tactile (T) feedback is becoming smaller while increasing the number of drones; meanwhile the relation between the T and V performance is increasing with increasing number of drones.

Comparing V and VT in Table 3, it is possible to conclude that VT, in general, showed slightly better average results than V. The best mean and absolute displacements for both experiments (best mean values: Experiment 1 – 9.5 mm, Experiment 2 – 18.3 mm) was also achieved with VT feedback and slow speed. Paired t-test showed no significant differences between V and VT in one drone landing, except V (slow landing case) and VT (fast landing case) case (t = 2.654, p = 0.012). Tactile feedback brings better performance to the right hand in the second experiment (comparing V and VT parameters in Table 3). For the right hand, visual plus tactile feedback is statistically better than only visual (V (fast landing case) and VT (slow landing case): t = 2.825, p = 0.008; V (fast landing case) and VT (fast landing case): t = 2.46, p = 0.019).

As a conclusion, we could state that the combination of visual and tactile feedback showed a synergetic effect.

TABLE III.  DRONE DISPLACEMENT, AFTER LANDING

| Feed-back type | Displacement statistics in XY plane, mm | | |
|---|---|---|---|
| | Mean, mm | Std. Deviation, mm | Maximum, mm |
| **Experiment 1. One drone. Right hand (Slow / fast landing)** | | | |
| V | 11.1/9.9 | 6.9/4.3 | 29.7/20.1 |
| T | 29.2/25.2 | 12.8/21.4 | 65.8/133 |
| VT | **9.5**/8.1 | 5.3/4.9 | 22.7/23.0 |
| **Experiment 2. Two drones. Left hand (Slow / fast landing)** | | | |
| V | 18.7/25.3 | 8.4/17.4 | 41.9/95.3 |
| T | 24.7/30.8 | 13.7/18.9 | 60.9/86.1 |
| VT | 20.7/22.2 | 11.5/13.4 | 45.5/57.3 |
| **Experiment 2. Two drones. Right hand (Slow / fast landing)** | | | |
| V | 31.4/19.2 | 23.1/14.5 | 116/58.8 |
| T | 28.7/51.1 | 14.6/113 | 63.5/143 |
| VT | **18.3**/20.9 | 11.0/14.4 | 47.1/63.6 |

Based on Fig. 5, in T condition of Experiment 2, the right hand demonstrated higher accuracy but low precision. However, the left hand was more precise and less accurate. Surprisingly for us, in terms of mean displacement and standard deviation, left hand worked out in a better way, for most participants. Left and right hands also performed in a different way with V feedback. Only in for VT, hands showed the same landing error parameters.

In general, performance with slow landing is better, which is obvious. Landing speed could strongly affect maximum error values, but on mean displacement and standard deviation it has a smaller effect in the most cases.

Hand motion patterns have been discussed previously in the trajectories analysis section. Using Fig. 5 we could support previous findings. For each feedback condition and each hand (for both experiments) we build a linear regression model with the least-squares approach, that predicts the position of drone landing. The results are presented with color lines in Fig. 5. It is possible to notice that the lines are always tilted from the center of the landing plate toward the human.

One of the most practical outcomes from the analysis of the positions after landing is the selection of a landing plate diameter. Diameter is the most important decision variable in the landing pad design. For the experiments, we selected such a size, that almost all landings were successful. As a result, now we could choose the percentage of successful landing that we want, and select the appropriate diameter. For example, for 90% of successful landings, the diameters are shown in Table 4 (shown in Fig. 5 as circles). If the drone lands not on the central part but on its legs, then the value has to be increased by the length of the leg.

TABLE IV. SELECTION OF THE LANDING PLATE DIAMETER

| Feed-back type | Diameter with 90% of successful landings, meters | | |
|---|---|---|---|
| | Experiment 1. Right hand. | Experiment 2. Left hand. | Experiment 2. Right hand. |
| V | 0.018 | 0.038 | 0.048 |
| T | 0.047 | 0.051 | 0.056 |
| VT | 0.017 | 0.039 | 0.039 |

The technology was demonstrated at ACM Siggraph Asia 2019 and won the Best Demonstration Award (honorable mentions) [13].

## 4 Conclusion and Future Work

We proposed a novel method and developed tactile interactive pads for the landing of the swarm of drones. During the experimental study, SwarmCloak demonstrated several significant advantages over pure visual feedback. It was shown that the tactile feedback allows the increasing accuracy of the landing pad positioning. It was also demonstrated that during the landing of two drones, tactile-visual feedback helped to considerably reduce the motion dynamics of the human head (snap is decreased by 14 times). Therefore, we can conclude that potentially the tactile channel reduces the stress of the operator. SwarmCloak is applicable when the vison feedback is not available, such as when users wear HMDs.

Two-way ANOVA of drone positions showed a statistically significant difference for all feedback/speed conditions. In contrast to the visual feedback, the number of drones does not significantly affect the performance of tactile feedback. The best landing positions were achieved with the combination of visual and tactile feedback. The paired t-test showed that for right-hand visual-tactile feedback is statistically better than only visual.

Although the current work considers landing of only two drones, a possible extension could be to arrange more landing pads on the forearms, upper arms to be able to land up to six drones on the operator body, which could require additional design development (the case of landing three drones on a hand is shown in Fig. 6). Regarding the limitation of the SwarmCloak, it could be hard to use the technology outdoors during the day time due to the not proper lighting conditions. Drones have to estimate the position of the human with an error less than a meter, which could be hard to do in some cases.

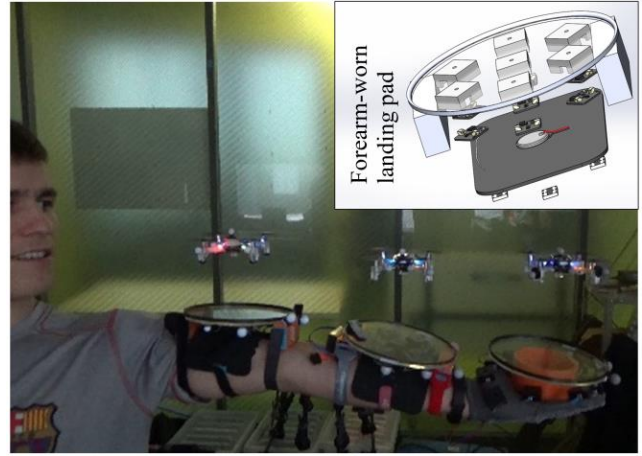

Fig. 6. Landing of three drones on one arm.

The proposed device can significantly augment the perception of flying/moving objects in Virtual Reality (VR) applications. Tactile sensations, such as bird landing or taking off from the human hands, can be simulated with SwarmCloak.